\documentclass[10pt,letterpaper]{article}
\usepackage[margin=0.72in,columnsep=0.25in]{geometry}
\usepackage[T1]{fontenc}
\usepackage{lmodern}
\usepackage{amsmath,amssymb,booktabs,graphicx,microtype,xcolor}
\usepackage[numbers,sort&compress]{natbib}
\usepackage[hidelinks,unicode]{hyperref}
\usepackage{xurl}
\usepackage{titlesec}
\titleformat{\section}{\large\bfseries\raggedright}{\thesection}{0.7em}{}
\titleformat{\subsection}{\normalsize\bfseries\raggedright}{\thesubsection}{0.7em}{}
\newcommand{\NLL}{\operatorname{NLL}}
\newcommand{\NCR}{\operatorname{NCR}}
\newcommand{\TQR}{\operatorname{TQR}}
\newcommand{\RGR}{\operatorname{RGR}}
\newcommand{\KL}{\operatorname{KL}}

\newcommand{\code}[1]{{\ttfamily\def\_{\char95\allowbreak}#1}}
\newcommand{\EOneGain}{0.747}
\newcommand{\ETwoGap}{0.076}
\newcommand{\ETwoNCR}{0.918}
\newcommand{\CorrectionParams}{434,176}
\hypersetup{pdftitle={LatentPort: Beyond KV Cache - Cross-Model Transfer of Recurrent Memory in Hybrid Language Models},pdfsubject={A 4B→9B Hybrid-State Handoff Without Target Prefix Replay},pdfauthor={Simon P. Villani}}

\begin{document}
\twocolumn[{
\begin{center}
{\LARGE\bfseries LatentPort: Beyond KV Cache -\\[3pt]
Cross-Model Transfer of Recurrent Memory\\[3pt]
in Hybrid Language Models\par}
\vspace{6pt}
{\large\itshape A 4B$\to$9B Hybrid-State Handoff Without Target Prefix Replay\par}
\vspace{10pt}
{\large Simon P. Villani}\\[13pt]
\end{center}
\begin{minipage}{\textwidth}
\begin{center}\textbf{Abstract}\end{center}
\small
Can one language model hand its live memory to another without the receiver rereading the context? We demonstrate useful persistent hybrid-state transfer across one architecture-matched Qwen3.5 4B$\to$9B sibling pair. To our knowledge, this is the first demonstrated cross-model handoff of persistent recurrent inference state between differently sized hybrid language models without target prefix replay. Translated attention KV alone leaves a large gap; adding the Gated DeltaNet (GDN) persistent-state package lowers teacher-forced negative log-likelihood (NLL), the average next-token log-loss, by \EOneGain{} nats/token (95\% paired document bootstrap CI [0.6921, 0.8047]), improving all 64 PG19 documents. Direct recurrent and convolution reuse outperforms the tested learned GDN maps, consistent with partial functional compatibility of persistent-state coordinates. A fresh component factorial selects translated KV with direct recurrent and convolution state. An additional rank-4 correction with \CorrectionParams{} trainable parameters improves that base on 64 fresh web documents: continuation loss is \ETwoGap{} nats/token above native 9B (excess NLL), Jensen--Shannon (JS) divergence is 0.022, and native context recovery (NCR) is \ETwoNCR{}. Corrected 9B significantly beats continued 4B inference while processing zero historical prefix tokens. Evidence covers one direction, one geometry-matched Base-model pair, and 4K teacher-forced continuation; the near-native gate failed, the 16K branch was not run, and free-generation equivalence and a general state interface remain unproven.
\end{minipage}
\vspace{16pt}
}]

\begin{figure*}[t]
\centering\includegraphics[width=\textwidth]{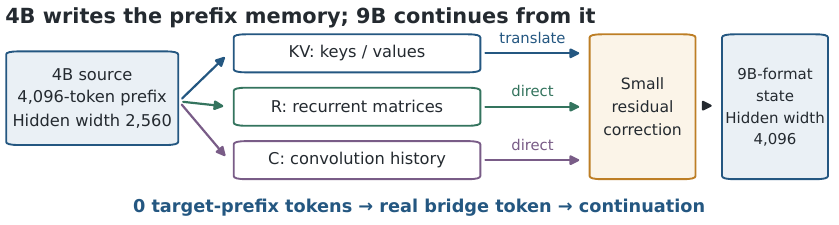}
\caption{\textbf{What crosses the model boundary.} The selected handoff translates attention KV, directly reuses GDN recurrent matrices and convolution history, and applies a small additional residual correction before installing target-compatible persistent state into the 9B runtime. The receiver processes a real bridge token and subsequent continuation, with \emph{zero historical prefix tokens}. Matched geometry permits copying despite different hidden widths and weights; usefulness is tested behaviorally. This schematic does not measure latency.}
\label{fig:schematic}
\end{figure*}

\begin{figure*}[t]
\centering
{\small\raggedright\textbf{Reading the figure.} Excess NLL is a condition's NLL minus native 9B NLL on the same continuation: 0 matches native continuation loss, and positive values are worse.\par}\vspace{5pt}
\includegraphics[width=\textwidth]{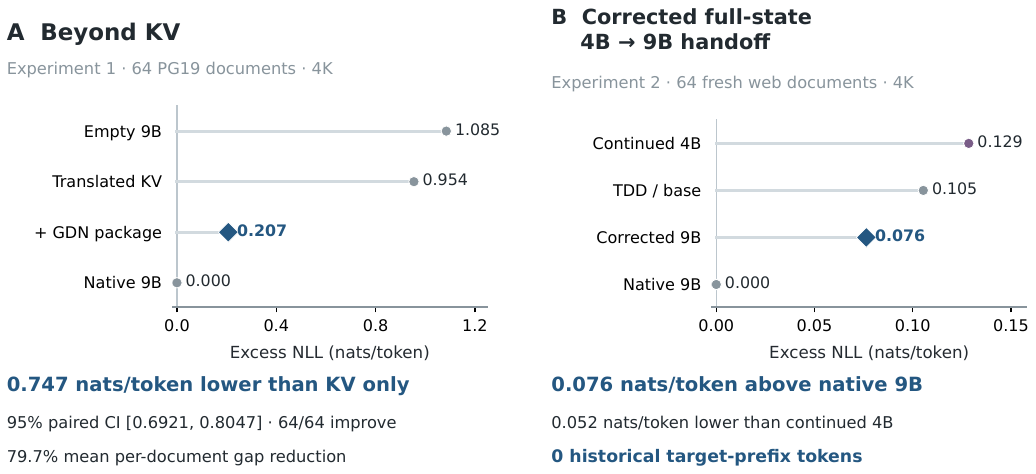}
\caption{\textbf{Beyond KV, then a corrected full-state handoff.} (A) Adding the full translated GDN persistent-state package to fixed translated KV improves all 64 paired PG19 documents, lowering NLL by 0.7473 nats/token (95\% CI [0.6921, 0.8047]). The mean per-document reduction of KV-only excess NLL is 79.7\%; recurrent matrices, convolution history, and initialization semantics are not isolated in this contrast. (B) On 64 different FineWeb-Edu test documents, the corrected handoff beats continued 4B inference by 0.0521 nats/token (95\% CI [0.0185, 0.0843]). The corrected handoff recovers 91.8\% of native 9B's prefix-derived NLL improvement relative to empty 9B; this is context-benefit recovery, not accuracy. Points are document means; intervals concern paired differences. Both experiments score 64 teacher-forced targets after a 4K prefix. Panel scales and corpora differ.}
\label{fig:hero}
\end{figure*}

\section{Introduction}
\label{sec:introduction}
Switching models usually requires the receiver to reread the historical prefix and rebuild inference state. Cross-model KV translation can avoid this repetition in full-attention models \citep{heo2026,qu2026,li2026}.

Hybrids retain more than KV: full-attention layers keep token-indexed keys and values, while Gated DeltaNet (GDN) layers keep recurrent matrices and convolution history \citep{yang2025}. KV-only transfer leaves this persistent memory behind.

Can that memory cross a model boundary? The source writes it with its own projections and gates; the receiver must read and update it with different weights. Matching shapes permit copying, but do not establish usefulness.

Qwen3.5-4B-Base and Qwen3.5-9B-Base differ substantially in parameter count and hidden width (2,560 versus 4,096), yet their GDN persistent-state geometry matches exactly. \emph{Direct recurrent and convolution reuse outperforms the tested learned GDN translation}. This is consistent with a partially shared functional coordinate system for persistent recurrent memory in the tested sibling pair.

Two frozen experiments measure teacher-forced negative log-likelihood (NLL): the average negative log-probability assigned to observed next tokens; lower is better. A nat is the natural-logarithm unit of information, so nats/token is average log-loss per predicted token. Adding the GDN package to translated KV lowers NLL by 0.747 nats/token, with all 64 test documents improving. Component selection and an additional 434,176-parameter correction yield a 9B handoff that significantly beats continued 4B, with zero historical target-prefix tokens (Figures~\ref{fig:schematic}--\ref{fig:hero}).

Evidence is limited to one directed, geometry-matched Base-model pair, 4K prefixes, and 64 teacher-forced targets per document. The near-native gate failed; no 16K, free-generation, downstream-task, or general-interface result follows.

\noindent\textbf{The specific first demonstration.}
Cross-model KV transfer, hidden-state messaging, external memory reuse, and same-model recurrent cache reuse are established directions (Section~\ref{sec:related}). Here, a differently sized hybrid receiver directly consumes persistent recurrent inference state written by another language model, without replaying the historical prefix. To our knowledge, prior work has not demonstrated cross-model transfer of built-in persistent recurrent inference state in an attention--recurrent hybrid language model without target prefix replay. Our contributions are:
\begin{itemize}
\setlength{\itemsep}{1pt}\setlength{\parskip}{0pt}
\item \textbf{Beyond-KV information.} A controlled intervention establishes a substantial GDN-package contribution beyond fixed translated KV.
\item \textbf{Functional compatibility.} Direct recurrent and convolution reuse beats tested learned maps across differently sized siblings.
\item \textbf{A constructive handoff.} Component selection and compact correction improve continuation over continued 4B without target prefix replay.
\item \textbf{Controlled evidence.} Paired document uncertainty, donor controls, frozen splits, and restoration checks support the result.
\end{itemize}

\section{Hybrid State and the Handoff Mechanism}
\label{sec:mechanism}
\subsection{Matched geometry, different models}
Both models repeat three GDN layers followed by one full-attention layer (Table~\ref{tab:architecture}). GDN combines gated forgetting with a delta-rule memory update \citep{yang2025}; in a conceptual value-by-key orientation,
\begin{align}
\bar S_t&=\alpha_t S_{t-1},\\
S_t&=\bar S_t+\beta_t(v_t-\bar S_t k_t)k_t^\top.
\end{align}
The runtime may store the transpose. Its persistent state also includes convolution history used to construct recurrent inputs. Both the accumulated matrix and these local histories can affect continuation.

\begin{table}[t]
\centering\small\setlength{\tabcolsep}{3pt}
\begin{tabular}{lrr}
\toprule
Property & 4B & 9B \\
\midrule
Parameters (billions) & 4.206 & 8.954 \\
Hidden width & 2560 & 4096 \\
Language layers & 32 & 32 \\
GDN / attention layers & 24 / 8 & 24 / 8 \\
Recurrent state / layer & $32\!\times\!128\!\times\!128$ & $32\!\times\!128\!\times\!128$ \\
Convolution state / layer & $8192\!\times\!4$ & $8192\!\times\!4$ \\
KV heads / head width & 4 / 256 & 4 / 256 \\
Attention query heads & 16 & 16 \\
\bottomrule
\end{tabular}

\caption{Runtime-verified correspondence. State shapes omit batch and are per GDN layer; each attention K or V tensor is $4\times L\times256$. Persistent geometry matches even though model size and hidden width differ.}
\label{tab:architecture}
\end{table}

Exact persistent geometry was a prerequisite for the experiment. Layer and head indices correspond directly; there is no cropping, padding, tiling, or learned index alignment. This correspondence makes direct installation mechanically valid but does not establish functional equivalence of states.

\subsection{What is installed and what is replayed}
After prefix $x_{1:L}$, write the source state as
\begin{equation}
\mathcal H_4(x_{1:L})=(K_4,R_4,C_4,m_4),
\end{equation}
where $K$ includes attention keys and values, $R$ the recurrent matrices, $C$ the convolution buffers, and $m$ the bookkeeping. Direct reuse or learned maps produce target-shaped components; deterministic runtime construction supplies metadata. Installation uses fresh cache objects and cloned storage.

The source reads $L=4096$ historical tokens. The receiver's first input is the real next token $x_{L+1}$, whose logits predict the first of 64 scored targets, $x_{L+2:L+65}$. This bridge is observed document text, not a learned prompt or historical replay. Every condition uses the same absolute positions and continuation schedule. Native target prefix processing supplies the baseline and offline training supervision; it is absent from the handoff path.

\paragraph{A package intervention, not recurrent matrices alone.}
KV-only leaves both GDN components fresh and uninitialized. Installing GDN state supplies recurrent matrices, convolution history, and correct initialization semantics together; initialized zero buffers would take a different runtime branch. The primary contrast therefore measures the \emph{GDN persistent-state package}. Its full effect cannot be attributed to recurrent matrices alone.

\subsection{Component translation and direct reuse}
Experiment 1 fits separate maps to paired source/target state. Attention K and V use ridge maps per layer, KV head, and role, with keys de-rotated before mapping and re-rotated for the target. For recurrent state, each layer/head uses
\begin{equation}
\widehat S_9=\mu_9+A(S_4-\mu_4)B^\top.
\label{eq:bilinear}
\end{equation}
Convolution buffers use grouped linear ridge maps. Regularization and permitted normalization are selected by validation tensor error, then frozen. Appendix~\ref{app:training} specifies the grid and positional inversion. Full translation maps $K,R,C$; the direct-GDN alternative uses the same translated $K$ and copies $R,C$.

Experiment 2 evaluates all eight direct/translated choices. A code lists $(K,R,C)$, with D for direct and T for translated. Validation selection uses mean excess NLL, a frozen 0.01-nat tie tolerance, then fewer translated components and lexical order. The selected TDD base translates KV and directly reuses both GDN components.

\subsection{An additional behavioral correction}
The correction is identity-anchored and leaves both language models frozen. For a KV feature matrix $X$,
\begin{equation}
X'=X+(XB)U^\top,
\label{eq:kvcorrection}
\end{equation}
where $B$ is fixed and orthonormal and $U$ is trainable, shared across heads and positions within a layer/role. Recurrent factors are shared across heads within a layer:
\begin{equation}
S'=S+U_LB_L^\top S+SB_RU_R^\top.
\label{eq:recurrentcorrection}
\end{equation}
Convolution uses $C'_{j,t}=C_{j,t}+a_jC_{j,t}+b_j$, shared over buffer positions. All trainable residual outputs start at zero; fixed bases remain nonzero. Components are optimized jointly through the receiver, without explicit cross-component tensor mixing.

The behavioral loss is full-vocabulary KL from native 9B to the handoff, averaged over nine output positions (bridge plus eight teacher inputs). An identity penalty averages squared residual/base norm ratios. Validation selects rank 4 and identity weight 0.01. The 434,176 trainable parameters are an \emph{additional correction}: they exclude the fitted KV translator, fixed bases, and both language models. They are not the size of the complete transfer system.

\section{Experimental Design}
\label{sec:design}
\subsection{Data and frozen selection}
Both experiments use official Base checkpoints in BF16, with FP32 recurrent storage, on an RTX 5090. Revisions, tokenizer hash, runtime versions, and selection rules appear in Appendices~\ref{app:runtime}--\ref{app:training}.

Experiment 1 calibrates translators on 128 FineWeb-Edu documents and validates on 32 disjoint documents, using 1,024-token prefixes and eight state checkpoints. FineWeb-Edu is an educational subset of FineWeb \citep{penedo2024}. Its held-out test set contains 64 PG19 books \citep{rae2020}, each with a 4,096-token prefix and 64 teacher-forced targets. This tests longer prefixes and a different corpus from calibration.

Experiment 2 uses 32 fresh FineWeb-Edu documents at 4K for the component factorial and 64 further fresh documents for the primary test. Both sets exclude all Experiment 1 split identities and text hashes. Correction training and validation reuse the earlier 128/32 calibration documents at 1K; no held-out outcomes enter fitting or selection. Frozen salted hash ordering, eligibility rules, and duplicate handling determine document membership independently of model outputs. The experiments' test corpora differ, so comparing their aggregate losses does not isolate a method improvement.

\subsection{Restoration and execution controls}
Before cross-model evaluation, complete same-model restoration is checked on 16 contexts per model across four lengths. Top-1 agreement is the fraction of positions where restored and native runs assign highest probability to the same next token; it is agreement, not accuracy. Both models achieve 100\% agreement and zero maximum absolute logit difference. This verifies capture and restoration, rather than state portability. A pre-fitting check also found chunked and one-shot prefix processing unequal; consequently, each calibration checkpoint uses an independent one-shot prefill. All compared conditions share the frozen continuation schedule (Appendix~\ref{app:runtime}).

Wrong-donor controls rotate complete states between equal-length documents with no self-donors. They test whether successful continuation depends on the appropriate prefix state. Neither experiment's conditional 16K branch ran. Experiment 2 additionally tracks state over 256 shared inputs; this is a secondary state diagnostic, not the primary 64-target quality endpoint.

\subsection{Metrics and uncertainty}
\label{sec:metrics}
Let $\ell_{d,c}$ be NLL over the 64 observed targets in document $d$ under condition $c$, using natural logarithms. Reported NLL is the equal-weight document mean $\ell_c=n^{-1}\sum_d\ell_{d,c}$, in nats/token. Excess NLL is the handoff NLL minus native 9B NLL on the same continuation:
\begin{equation}
\Delta\NLL_c=\ell_c-\ell_{\mathrm{native9B}}.
\end{equation}
An excess NLL of 0 matches native continuation loss; positive values are worse, and negative values are lower loss. Equal loss does not imply identical predictions. We also compare directly with continued source 4B.

Jensen--Shannon (JS) divergence measures how different the full output probability distribution is from native 9B: 0 means identical distributions and lower is better. We average it over scored positions and documents. Top-1 agreement compares only the highest-probability next token, using native 9B as the reference.

Native context recovery (NCR) measures how much of the continuation benefit that native 9B obtains from processing the prefix is recovered by the handoff, relative to an empty 9B state:
\begin{equation}
\NCR_c=\frac{\ell_{\mathrm{empty9B}}-\ell_c}
{\ell_{\mathrm{empty9B}}-\ell_{\mathrm{native9B}}}.
\label{eq:ncr}
\end{equation}
Thus, NCR $=0.918$ means recovering 91.8\% of native 9B's improvement in continuation NLL from having the prefix, relative to empty 9B. It does \emph{not} mean 91.8\% accuracy. NCR is a ratio of aggregate NLLs and can lie outside $[0,1]$. Appendix~\ref{app:metrics} defines target-quality recovery (TQR), remaining-gap reduction, and distribution/state diagnostics.

The statistical unit is the document. Intervals use 10,000 paired document bootstrap resamples and central 95\% percentiles with frozen seeds. They preserve within-document condition pairing; tokens are not independent replicates. Verification reproduces the frozen calculations and retains canonical intervals. Post-verdict diagnostics are explicitly separated from primary results.

\begin{center}
\begin{minipage}{\columnwidth}
\small\centering\setlength{\tabcolsep}{3pt}
\begin{tabular}{@{}p{.25\columnwidth}p{.48\columnwidth}p{.21\columnwidth}@{}}
\toprule
Metric & Meaning & Reference \\
\midrule
\raggedright NLL & \raggedright Average log-loss on observed continuation & Lower \\
\raggedright Excess NLL & \raggedright Handoff NLL minus native 9B NLL & 0 \\
\raggedright NCR & \raggedright Fraction of native 9B context benefit recovered & 1 \\
\raggedright JS divergence & \raggedright Difference from native 9B output distribution & 0 \\
\raggedright Top-1 agreement & \raggedright Same highest-probability next token as native 9B & Higher \\
\bottomrule
\end{tabular}

\makeatletter\def\@captype{table}\makeatother
\caption{\textbf{Metrics at a glance.} Reference values describe loss or agreement with native 9B, not task accuracy.}
\label{tab:metricguide}
\end{minipage}
\end{center}

\section{Results}
\subsection{Does GDN memory contribute beyond KV?}
\label{sec:beyondkv}
\begin{figure}[t]
\centering\includegraphics[width=\columnwidth]{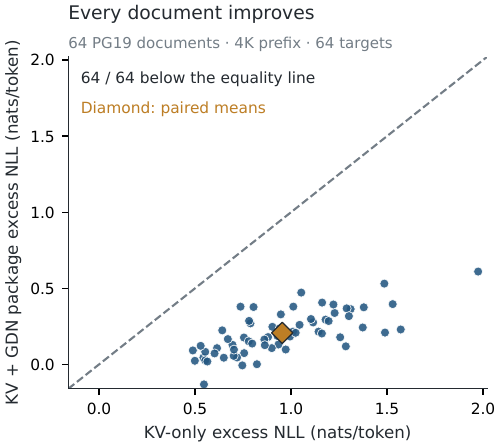}
\caption{\textbf{64 of 64 documents improve.} Each point pairs one PG19 document's excess NLL under KV-only and full translated state; every point lies below equality. Adding the GDN package improves NLL by 0.7473 nats/token (95\% paired document bootstrap CI [0.6921, 0.8047]). The diamond marks paired means. Prefixes contain 4,096 tokens and each document contributes 64 teacher-forced targets. This is a document-level effect, not 4,096 independent token replicates.}
\label{fig:e001}
\end{figure}

Holding translated KV fixed, adding the translated GDN persistent-state package lowers NLL by 0.7473 nats/token (95\% CI [0.6921, 0.8047]). \emph{All 64 documents improve} (Figure~\ref{fig:e001}). The direction is therefore consistent across the entire test set, rather than confined to a small subset of documents. Each scatter point is one paired document, the same unit used for uncertainty.

\begin{table}[t]
\centering\small\setlength{\tabcolsep}{3.5pt}
\begin{tabular}{lrrr}
\toprule
Condition & NLL & $\Delta$NLL & Top-1 \\
\midrule
Native 9B & 2.1609 & 0.0000 & 1.000 \\
Source 4B & 2.3074 & 0.1465 & -- \\
Empty 9B & 3.2457 & 1.0847 & 0.581 \\
KV only & 3.1153 & 0.9544 & 0.600 \\
KV + direct GDN & 2.2913 & 0.1304 & 0.821 \\
Full translated & 2.3679 & 0.2070 & 0.783 \\
Full shuffled & 3.3162 & 1.1553 & 0.556 \\
\bottomrule
\end{tabular}

\caption{Experiment 1 test fidelity on 64 PG19 documents. NLLs are document means in nats/token; top-1 is agreement with native 9B. Source agreement was not recorded in this table. Full precision is retained in Appendix~\ref{app:fullresults}.}
\label{tab:e001}
\end{table}

KV-only NLL is 3.115, close to empty 9B's 3.246 and far above native 9B's 2.161 (Table~\ref{tab:e001}). Adding GDN state lowers it to 2.368 and removes about 80\% of KV-only excess NLL: the mean per-document reduction is 79.7\% and the median 80.2\%. This package includes recurrent matrices, convolution history, and initialization semantics. The primary contrast does not separate their individual contributions.

The fully translated condition nevertheless remains worse than continued 4B (2.307 NLL). It clears the preregistered recurrent-state contribution criterion but misses the stronger full-state criterion, with excess NLL 0.207. The exact experiment-specific verdicts and gates are retained in Appendix~\ref{app:metrics}.

\subsection{Which components need translation?}
\label{sec:components}
\begin{figure*}[t]
\centering\includegraphics[width=\textwidth]{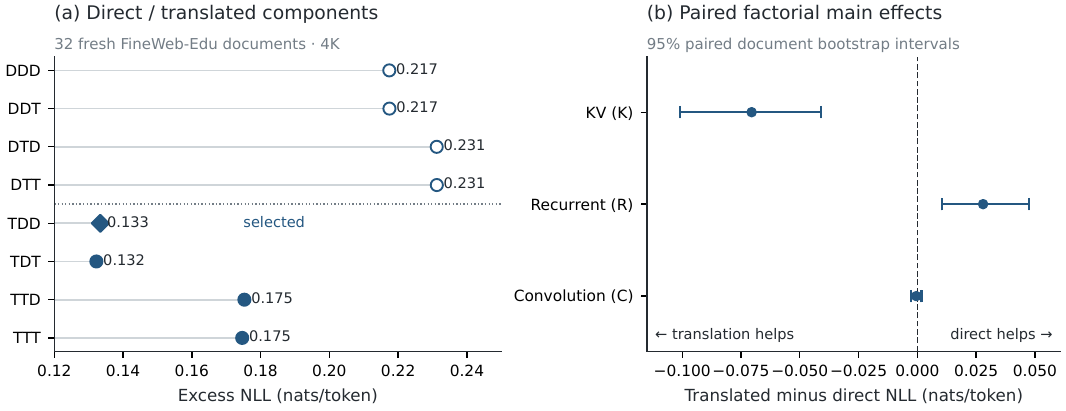}
\caption{\textbf{Translate KV; directly reuse recurrent memory.} Experiment 2's factorial uses 32 fresh FineWeb-Edu validation documents, 4K prefixes, and 64 teacher-forced targets. (A) D/T denotes direct/translated $(K,R,C)$ state; filled points translate KV. TDT has the lowest numerical loss, but the frozen 0.01-nat tie rule selects TDD (diamond), which uses fewer translations. (B) Main effects average paired differences over the other factors; bars are 95\% document bootstrap intervals. Recurrent translation worsens NLL, whereas convolution translation has an uncertain small effect.}
\label{fig:factorial}
\end{figure*}

Direct GDN reuse is stronger than the tested learned GDN maps. In Experiment 1, replacing recurrent and convolution translation with direct copying reduces NLL from 2.368 to 2.291, a 0.077-nat improvement with translated KV unchanged. Experiment 2 tests this component preference afresh (Figure~\ref{fig:factorial}). The learned recurrent mapper nevertheless has lower validation reconstruction error; the behavioral comparison also changes convolution translation and evaluation domain (Section~\ref{sec:reconstruction}).

In the 32-document factorial, translating KV lowers NLL by 0.070 nats/token on average (translated minus direct: $-0.0705$, CI [$-0.1009$, $-0.0409$]). Translating recurrent state instead raises NLL by 0.0280 (CI [0.0104, 0.0475]). Convolution translation's estimate is $-0.0004$ (CI [$-0.0026$, 0.0018]); this interval does not establish equivalence.

TDT has the lowest validation excess NLL, 0.132226, versus 0.133377 for TDD. The difference falls within the frozen tie tolerance, selecting TDD because it translates fewer components. On the separate 64-document test set, TDD reaches excess NLL 0.105, compared with 0.134 for TTT (paired TTT-minus-base CI [0.0170, 0.0418]). Direct GDN preference thus persists on held-out continuations.

No pairwise interaction satisfies the preregistered criterion for replication in both direction and magnitude (Appendix~\ref{app:factorial}). The validation KV--recurrent interaction is 0.0284 nats/token; its test estimate, 0.0151, is below the 0.02 materiality threshold. Joint optimization of a correction therefore need not imply that material cross-component coupling explains the original mismatch.

\subsection{Can the corrected handoff beat continued 4B?}
\label{sec:correctionresult}
Yes, on the held-out teacher-forced continuations. Corrected 9B NLL is 1.989 versus 2.042 for continued 4B. Corrected minus source is $-0.0521$ nats/token (95\% CI [$-0.0843$, $-0.0185$]). This supplies an intuitive behavioral endpoint: after receiving the imported state, the larger model predicts the observed continuation better than the source that read the prefix.

\begin{table*}[t]
\centering\small\setlength{\tabcolsep}{9pt}
\begin{tabular}{lrrrrr}
\toprule
Condition & NLL & $\Delta$NLL & JS & NCR & Top-1 \\
\midrule
Native 9B & 1.913 & 0.000 & 0.000 & 1.000 & 1.000 \\
Source 4B & 2.042 & 0.129 & 0.038 & 0.862 & 0.834 \\
Empty 9B & 2.842 & 0.929 & 0.157 & 0.000 & 0.626 \\
TDD / base & 2.018 & 0.105 & 0.029 & 0.886 & 0.846 \\
TTT & 2.047 & 0.134 & 0.036 & 0.855 & 0.832 \\
Joint corrected & 1.989 & 0.076 & 0.022 & 0.918 & 0.864 \\
Joint shuffled & 3.178 & 1.265 & 0.192 & -0.362 & 0.578 \\
\bottomrule
\end{tabular}

\caption{Experiment 2 test fidelity on 64 fresh FineWeb-Edu documents at 4K, with 64 teacher-forced targets each. Corrected 9B improves on both its selected base and continued 4B; wrong-donor state removes the advantage. NLL and excess NLL are nats/token, JS compares distributions with native 9B, and top-1 measures prediction agreement. NCR is context-benefit recovery (Eq.~\ref{eq:ncr}). Six-decimal values and TQR appear in Appendix~\ref{app:fullresults}.}
\label{tab:e002}
\end{table*}

The correction lowers base NLL from 2.018 to 1.989 and excess NLL from 0.105 to 0.076 (Table~\ref{tab:e002}). The paired improvement is 0.0290 nats/token (CI [0.0228, 0.0354]), removing 27.5\% of the remaining native gap (ratio CI [22.4\%, 33.9\%]). The corrected continuation loss is only 0.076 nats/token above native 9B. Its full output distribution is also close to native 9B (JS 0.022, where 0 is identical), improving from base JS 0.029. NCR 0.918 corresponds to recovering 91.8\% of native 9B's prefix-derived continuation benefit relative to empty 9B. Top-1 agreement rises from 0.846 to 0.864.

These gains require only a small adjustment to the base: the median layer-relative residual norm is 0.0264, and the maximum is 0.0440. The 434,176-parameter correction occupies 1.90 MB serialized, in addition to the existing KV translator (Appendix~\ref{app:training}). These norms are relative to the base state, not bounds on prediction error.

The result passes the protocol's full-state gate, but \emph{fails its near-native gate}: excess NLL 0.076 exceeds 0.05, NCR 0.918 is below 0.95, and top-1 agreement 0.864 is below 0.90. All three stronger thresholds are missed; the conditional 16K branch was therefore not run. Appendix~\ref{app:metrics} records the exact \code{FULL\_STATE\_HANDOFF} and \code{NEAR\_NATIVE\_HANDOFF} definitions. The measured gain is teacher-forced fidelity, not free-generation or downstream-task equivalence.

\subsection{Does the correct source context matter?}
\label{sec:donor}
Rotating complete donor states removes the continuation advantage. In Experiment 1, correct full state improves on shuffled state by 0.948 nats/token (CI [0.878, 1.022]). In Experiment 2, corrected minus shuffled NLL is $-1.188$ (CI [$-1.296$, $-1.085$]); shuffled NLL is 3.178, worse than empty 9B's 2.842.

The receiver therefore uses document-specific information in the transferred hybrid state. Together with the fixed-KV primary contrast, these controls support useful transfer beyond attention KV. Because the shuffle changes the complete donor state, it does not independently localize that context specificity to recurrent matrices or identify particular retained facts. Secondary state-convergence measurements and explicitly post-verdict correction removals are reported in Appendix~\ref{app:diagnostics}.

\section{Related Work: Interfaces for Reuse}
\label{sec:related}
\subsection{Cross-model KV/cache translation}
Cache-to-Cache projects and fuses source KV with the receiver's own context cache \citep{fu2026}. Latent Cache Flow adds joint K/V bottlenecks and pooled communication across differing contexts, retaining receiver-cache fusion \citep{rossi2026}. Semantic Cache Distillation reconstructs KV from compact codes and sparse normalized hidden-input patches for shared-architecture, weight-mismatched models \citep{ma2026}. It transfers more than raw KV but does not evaluate persistent GDN state. Mixture-of-Translators maps heterogeneous KV and uses target context replay to reconstruct cache, with source-guided sparsification \citep{lee2026}.

XKV pools both models' KV caches into joint cross-layer memory and lets each receiver position retrieve a gated KV residual \citep{liu2026xkv}. It supports heterogeneous models holding complementary private contexts. Both models prefill their own contexts; the communication updates attention KV, rather than transferring built-in persistent recurrent state.

Other work installs translated KV without full receiver prefix processing. \citet{heo2026} study within-family transfer in dense full-attention models using ridge maps, layer selection, and RoPE factoring; attention--recurrent hybrids are outside their evaluation. CacheBridge uses head-local support and attention-weighted calibration \citep{qu2026}. A Universal Context-Reuse Layer reports KV sharing within and across families \citep{li2026}. KV transfer itself is established prior work. Our additional question concerns persistent recurrent and convolution state in a live hybrid receiver.

\subsection{Hidden-state and latent communication}
StateBridge aligns message hidden states to an embedding interface and supplies a continuous prefix processed by the receiver \citep{peng2026}; its reported multi-agent runs share weights within each run. A continuous message can communicate useful information without installing state in the receiver's built-in persistent inference slots. We distinguish these interfaces without comparing unlike tasks or replay budgets.

\subsection{External memory and reader adaptation}
\citet{li2026reader} transfer learned Engram-style external memory through a tokenizer-agnostic addressing interface. They study both direct reuse of compatible memory/reader artifacts and target-side reader adaptation with frozen memory and backbones. Their result makes reader compatibility directly relevant to our question. The transferred object, however, is a learned external memory with injected reader outputs, rather than a hybrid model's built-in inference cache captured after a particular input prefix. Our handoff installs that input-conditioned persistent state without adapting target backbone weights.

\subsection{Cross-model activation-state transfer}
\citet{piepereit2026} project intermediate activations across model architectures and test their effect through activation injection. Alignment scores do not consistently predict successful behavioral transfer, and the observed effects depend on the model pair. Their generation intervention replaces a prompt-position hidden state while reprocessing token sequences; it does not install persistent hybrid cache state. This is adjacent evidence about functional compatibility of internal representations, rather than evidence that all internal state is portable.

\subsection{Same-model recurrent/hybrid reuse}
Marconi manages hybrid prefix checkpoints for serving \citep{pan2025}. HYPIC composes recurrent state from segment operators and end-states, with seam recomputation for hybrid attention \citep{liu2026hypic}. DASC selects and compresses persistent state, using zero filling or bounded refresh on restoration \citep{yu2026}. Tail-Replay reconstructs recurrent state from a recent suffix of a matched prefix instead of retaining recurrent checkpoints \citep{liu2026tail}. These systems reuse, compose, or recover memory for the same model. Here, a different model consumes the state and performs no historical prefix replay.

WriteSAE decomposes recurrent cache state into matrix atoms and tests replacement or amplification of writes in Gated DeltaNet and related architectures \citep{young2026writesae}. These behavioral interventions establish recurrent cache writes as a causal intervention surface. Its larger-host probe also reuses a learned dictionary without retraining. That transferred dictionary edits the host's state; it is not a prefix-conditioned recurrent state written by another model and installed for receiver continuation.

Recurrent-state initialization is also an adaptation surface: S0 tuning learns initial matrices before processing a prompt, rather than transferring another model's prefix-conditioned state \citep{young2026}. Metis adds learned memory blocks and evaluates memory-only inference on multiple backbones \citep{zhang2026metis}. Its backbone-transfer study trains the memory mechanism for each backbone; it does not demonstrate a live state written by one model being consumed by another.

\subsection{Positioning}
\label{sec:positioning}
We are not aware of prior work demonstrating cross-model transfer of built-in persistent recurrent inference state between differently sized attention--recurrent hybrid LLMs without target prefix replay. Here, \emph{built-in persistent} identifies the model's built-in recurrent inference interface; imported state is target-compatible state, not the state native 9B would produce by reading the prefix. The distinction is the combination of different models and sizes, hybrid recurrence, prefix-conditioned persistent state, live receiver installation, and zero historical receiver replay. Our literature check supports this qualified claim, not a claim that cross-model internal-state transfer generally is new.

\section{Discussion}
\paragraph{Partial functional state compatibility.}
Equal shapes explain why copying is possible, not why it works. The two models have different hidden widths and learned read, write, and gating operations. Direct GDN reuse nevertheless supports strong continuation when paired with translated KV, and a wrong donor sharply degrades performance. This is consistent with partially compatible functional coordinates for persistent state in this sibling pair. Shared pretraining or a shared training recipe could help produce such compatibility; that is a hypothesis, not a factor isolated here. Neither representation identity nor a formal state ABI follows from the result.

\paragraph{Lower state reconstruction error does not guarantee better cross-model continuation.}
\label{sec:reconstruction}
The learned recurrent mapper reduces normalized recurrent-state reconstruction error on validation data from 0.6732 for direct copying to 0.5037, yet full learned GDN translation performs worse behaviorally than direct GDN reuse. In this experiment, lower recurrent tensor reconstruction error is therefore insufficient to predict a better handoff. The behavioral contrast also changes convolution translation, and calibration and held-out continuation use different domains. These confounds prevent an isolated causal conclusion about recurrent-state Euclidean error. The empirical lesson is consistent with CacheBridge: attention-sensitive weighting improves continuation despite little change in unweighted KV reconstruction \citep{qu2026}. A behavioral objective directly targets receiver compatibility. Post-verdict removals identify recurrent correction as influential in the fitted solution, but do not isolate the recurrent matrix's share of the original package benefit (Appendix~\ref{app:ablations}).

\paragraph{Bounded portability beyond KV.}
The result is an existence demonstration under favorable correspondence: persistent recurrent and convolution memory can remain useful after a change of model. It does not supply a universal translation rule. Geometry mismatch, altered head/layer organization, or different learned update dynamics could break the observed compatibility. Prototype timings in Appendix~\ref{app:timing} do not establish serving speedup.

\paragraph{A possible state-interface design direction.}
Direct reuse motivates a speculative design direction: model families could be trained deliberately to preserve compatible persistent-state geometry and functional coordinates across sizes, exposing a stable interface for model switching. Such an interface could function as a state ABI between model sizes. This experiment establishes neither that such an ABI exists generally nor that it can be engineered; it only motivates testing the idea.

\paragraph{Decisive next tests.}
The two most decisive next tests are unconstrained free generation after handoff and replication on an independent model pair. Free generation tests whether small state errors compound when ground-truth continuation tokens no longer stabilize the receiver. An independent pair tests whether compatibility is peculiar to these Qwen siblings or reproducible in families with matched persistent-state geometry. Mismatched geometry is a subsequent, harder test. These are follow-up research directions, not completed evidence.

\section{Limitations}
The study covers one directed 4B$\to$9B transfer between Qwen3.5 Base siblings with exactly matched persistent-state geometry. Translators and correction are specific to this pair. Instruct models, cross-family transfer, and mismatched geometry are untested; shared pretraining and training recipe may contribute to the result.

Behavioral evaluation uses 4K prefixes and 64 teacher-forced targets. Ground-truth inputs can stabilize continuation, so neither free-generation equivalence nor downstream task success follows. The 16K branch did not run; the 256-input state diagnostic is not a longer-context quality evaluation. The original primary contrast does not isolate recurrent matrices from convolution history and initialization semantics.

Runtime scheduling matters, as the chunk-equivalence check demonstrates. Bootstrap uncertainty is conditional on the frozen models, fitted maps, selected correction, and corpus procedure; it does not cover training-seed or model-pair variation. Post-verdict ablations reuse test documents and are exploratory. Timing omits important costs and is not an end-to-end production benchmark.

\section{Conclusion}
\label{sec:conclusion}
KV is not the whole transferable inference state of a hybrid LLM. In one matched Qwen3.5 4B$\to$9B pair, persistent GDN state carries source-specific information that the larger receiver can use without reading the historical prefix. To our knowledge, this is the first demonstrated cross-model handoff of persistent recurrent inference state between differently sized hybrid language models without target prefix replay. Direct recurrent and convolution reuse is unexpectedly strong despite the learned recurrent mapper's lower reconstruction error, suggesting partial functional compatibility of persistent-state coordinates; a compact additional correction further closes the continuation gap. This bounded result motivates deliberately training families for persistent-state compatibility across sizes, an idea that requires free-generation testing and replication beyond this sibling pair.

\begingroup\small
\bibliographystyle{plainnat}
\bibliography{references}
\endgroup
\clearpage
\appendix
\setcounter{dbltopnumber}{3}
\small
\section{Frozen Models and Runtime}
\label{app:runtime}
The source repository is \code{Qwen/Qwen3.5-4B-Base}, revision
\begin{quote}\footnotesize\path{1001bb4d826a52d1f399e183466143f4da7b741b}.\end{quote}
The target is \code{Qwen/Qwen3.5-9B-Base}, revision
\begin{quote}\footnotesize\path{68c46c4b3498877f3ef123c856ecfde50c39f404}.\end{quote}
The tokenizer uses the source repository and revision. The frozen \code{tokenizer.json} SHA-256 is
\begin{quote}\footnotesize\path{fe000e3ed39ed12b8d2481d527d44f93c65d37e87645d2dcc80d1bf9d50d2927}.\end{quote}
These are official Base checkpoints. No instruct substitution, weight updates, LoRA, quantization, MTP, or speculative decoding is part of either experiment. Loaded language-model parameter counts are 4,205,751,296 and 8,953,803,264.

The recorded runtime uses Python 3.11.9, PyTorch 2.13.0+cu130, CUDA runtime 13.0, Transformers 5.16.1, Datasets 5.0.1, and Hugging Face Hub 1.29.0. Hardware is an NVIDIA GeForce RTX 5090, compute capability 12.0, with 32,607 MiB reported VRAM; driver 591.86 reports CUDA 13.1 support. These version numbers are taken from the frozen runtime artifact. They describe the experimental environment, not the lighter Python/LaTeX environment used to render this paper.

The architecture gate checks all 32 layer types, attention KV geometry, the GDN recurrent and convolution geometry, and RoPE geometry. Attention has eight layers at zero-based indices $3,7,\ldots,31$. Each K or V tensor is $1\times4\times L\times256$. Each of the other 24 layers retains one recurrent tensor of shape $1\times32\times128\times128$ and a convolution tensor of shape $1\times8192\times4$. Recurrent memory is FP32; attention and convolution memory use BF16. At 4K, the complete tensor payload is 186,122,240 bytes for either model. Metadata includes initialization flags, previous-state flags, convolution kernel sizes, logical length, and positions; it is constructed deterministically rather than learned.

Complete same-model restoration uses 16 deterministic contexts per model at lengths 256, 512, 1,024, and 2,048. Both achieve top-1 agreement 1.0 and maximum absolute logit difference 0; the minimum next-logit cosine is 0.99999988. A separate pre-fitting one-shot versus eight-chunk 1,024-token check has maximum bridge-logit difference 0.125. Each FIT/VALIDATION checkpoint was therefore captured by an independent fresh-cache one-shot prefill at 128-token intervals. Canonical continuation shares the 1/4/16/64 checkpoint schedule across conditions; E002 extends state tracking to 256 inputs. These are frozen implementation decisions.

The installer rejects activation of only one GDN component: recurrent and convolution restoration share a previous-state branch. Consequently, an independently fresh-R/direct-C or direct-R/fresh-C intervention is not implemented by the frozen path. Bypassing the guard and supplying initialized zeros would change the meaning of ``fresh.'' No new presence ablation was run for this revision; isolating R from C requires a separately validated intervention. The local revision feasibility audit records the exact code paths. This does not alter either original experiment.

\section{Splits and Deterministic Selection}
\label{app:data}
E001 uses FineWeb-Edu revision
\begin{quote}\footnotesize\path{87f09149ef4734204d70ed1d046ddc9ca3f2b8f9}\end{quote}
and immutable shard \path{sample/100BT/000_00000.parquet}. Its FIT/VALIDATION candidate universe is physical rows 0--19,999. Eligible rows are sorted by a SHA-256 key containing the frozen salt \code{latentport-e001-splits-v1}, corpus revision, shard, physical row, and document ID. Duplicate document IDs or UTF-8 text hashes retain the lowest-ranked row. The first 128 eligible documents form FIT and the next 32 VALIDATION, each requiring at least 1,089 token IDs.

PG19 uses \code{emozilla/pg19}, revision
\begin{quote}\footnotesize\path{c021754c8e01c5b1cc83a1f549c1f97fbbb756b8},\end{quote}
and its frozen test parquet. Hash ordering first reserves 16 documents meeting the 16,449-token LONG requirement; LOCKED takes the first 64 remaining eligible documents meeting the 4,161-token requirement. Reserve membership is part of the frozen selection protocol, not evidence of LONG evaluation.

E002 uses the same FineWeb-Edu revision and shard but expands the candidate universe to rows 0--99,999 and uses the salt \code{latentport-e002-splits-v1}. Its coarse filter uses source-reported token count before exact eligibility under the frozen Qwen tokenizer. All E001 split IDs and text hashes are excluded from fresh selection. The factorial reserves 32 fresh 4K documents; LOCKED contains 64 further fresh 4K documents. The LOCKED records store 4,352 tokens to support bridge-plus-255-input state tracking. Correction fitting/validation explicitly reference the 128/32 E001 non-LOCKED documents. Split IDs, source hashes, and input tokens remain frozen. The paper verification reads no unused LONG token payloads.

\section{Translator and Correction Details}
\label{app:training}
The E001 ridge grid is $\{10^{-8},10^{-6},10^{-4},10^{-2},1,10^2\}$ relative to $\operatorname{tr}(X_c^\top X_c)/d$. KV and convolution fits consider centering alone or per-feature standardization; recurrent fits use centering and exactly three alternating updates of Eq.~\ref{eq:bilinear}. The KV position sampler selects 64 positions per FIT document, including endpoints and 128-token boundaries. Recurrent fitting uses eight independent one-shot checkpoint captures per document. Hyperparameter selection minimizes registered validation tensor error; no layer/head alignment or position parameters are learned.

The frozen E001 operator weights comprise 4,194,304 KV coefficients, 25,165,824 recurrent coefficients, and 25,165,824 convolution coefficients: 54,525,952 in total, excluding stored centering/normalization values. The full bundle occupies 322,222,408 bytes. E002's selected TDD path uses the KV operator and direct GDN state. Its KV tensor artifact alone is 17,042,160 bytes, in addition to the 1,901,432-byte correction. Thus correction size should not be confused with total deployed mapping storage.

E002 uses rank $r\in\{2,4\}$ and identity penalty $\lambda_{\rm id}\in\{0,10^{-4},10^{-3},10^{-2}\}$. Table~\ref{tab:traininggrid} reports every selected-per-candidate checkpoint. The fixed Gaussian-QR basis seed is 2026090102; the global/training seeds are 2026090101 and 2026090106. AdamW uses learning rate $10^{-3}$, $(\beta_1,\beta_2)=(0.9,0.95)$, $\epsilon=10^{-8}$, no weight decay, one document per batch, gradient accumulation over eight documents, and gradient-norm clipping at 1. Training follows manifest order for at most three epochs, with early-stop patience one and minimum improvement $10^{-6}$. Candidate selection minimizes validation behavioral KL; ties within $10^{-6}$ prefer lower rank, larger identity penalty, then earlier epoch. The selected checkpoint has rank 4, $\lambda_{\rm id}=0.01$, and epoch 3.

\begin{table}[t]
\centering\small\begin{tabular}{lrrr}
\toprule
Rank & $\lambda_{\rm id}$ & Epoch & Val. KL \\
\midrule
2 & 0 & 3 & 0.171111 \\
2 & 0.0001 & 3 & 0.169789 \\
2 & 0.001 & 3 & 0.170405 \\
2 & 0.01 & 3 & 0.170945 \\
4 & 0 & 3 & 0.162947 \\
4 & 0.0001 & 3 & 0.163098 \\
4 & 0.001 & 3 & 0.162811 \\
4 & 0.01 & 3 & 0.161918 \\
\bottomrule
\end{tabular}

\caption{Frozen E002 correction selection grid. Validation KL is mean full-vocabulary native-to-handoff KL over nine output positions on 32 non-LOCKED 1K documents. These are selection scores, not LOCKED results.}
\label{tab:traininggrid}
\end{table}

The rank-4 trainable count is
\begin{align*}
P_K&=8\cdot2\cdot256\cdot4=16{,}384,\\
P_R&=24\cdot2\cdot128\cdot4=24{,}576,\\
P_C&=24\cdot2\cdot8192=393{,}216.
\end{align*}
The fixed bases are buffers, not trainable parameters. All trainable outputs, scales, and biases start at zero. The layer-relative magnitude statistics pool 64 documents' records; they are not just norms of the trained weights. KV norms combine the K and V residuals within each attention layer, while recurrent and convolution records are separate. The pooled median therefore reflects this explicit record weighting.

Attention translation fits each layer, KV head, and K/V role separately. Its positional step operates on the 64 rotary dimensions of each 256-dimensional key. The implementation explicitly inverts the BF16-rounded sine/cosine transform, dividing by the corresponding squared-sine-plus-squared-cosine factor; nonrotary coordinates are unchanged by that step. Recurrent bilinear maps use three deterministic alternating ridge updates per regularization candidate. Convolution maps operate within the runtime's packed Q/K/V component and head groups. Selection uses validation tensor error before either held-out evaluation.

The selected correction contains 434,176 trainable parameters under an exclusive two-million-parameter cap. Its serialized artifact is 1,901,432 bytes, $1.0618013\times10^{-4}$ of target BF16 parameter bytes. The pooled median residual/base norm is 0.026432 and the maximum 0.043951. Table~\ref{tab:correction} gives the component inventory. These magnitudes concern the installed state perturbation, not native-state reconstruction error or prediction error.

\begin{table}[t]
\centering\small\setlength{\tabcolsep}{4pt}\begin{tabular}{lrr}
\toprule
Component & Parameters & Median norm \\
\midrule
KV (rank 4) & 16,384 & 0.027093 \\
Recurrent (rank 4) & 24,576 & 0.031916 \\
Conv. scale / bias & 393,216 & 0.015320 \\
Total / pooled & 434,176 & 0.026432 \\
\bottomrule
\end{tabular}

\caption{Frozen E002 correction inventory. Median norm is the residual/base Frobenius norm over LOCKED document--layer records, pooled separately by component or over all records. Rank 4 applies to KV and recurrent residuals; convolution uses channel scale/bias. Counts exclude fixed bases and the existing KV translator.}
\label{tab:correction}
\end{table}

\section{Condition and Metric Semantics}
\label{app:metrics}
Native 9B processes the historical prefix normally; source 4B continues from its own prefix state. Empty 9B begins without historical state at the same logical continuation position. E001 KV-only installs translated attention KV and leaves both GDN components fresh. KV plus direct GDN installs translated KV with copied recurrent and convolution tensors. Full translated installs all three frozen mapped components. Full shuffled uses a different document's complete translated state. E002 D/T cells install the stated source or translated components, base is frozen TDD, corrected applies Eqs.~\ref{eq:kvcorrection}--\ref{eq:recurrentcorrection} and the convolution residual, and joint shuffled rotates complete corrected donor states. Donor rotations are within equal-length records, with no self-donors.

The bridge is input token $x_{L+1}$; 64 scored targets are $x_{L+2:L+65}$. The segmented canonical path consumes the bridge and the first 63 continuation targets as inputs. A state checkpoint labeled 64 therefore follows 64 post-handoff inputs, including the bridge. E002 training uses the bridge plus eight teacher inputs to supply nine output distributions, in one differentiable target call. Native target traces needed for fitting are offline supervision; they do not imply target historical replay in the deployed handoff.

For protocol completeness, target-quality recovery and remaining-gap reduction are
\begin{align}
\TQR_c&=\frac{\ell_{\mathrm{source4B}}-\ell_c}{\ell_{\mathrm{source4B}}-\ell_{\mathrm{native9B}}},\label{eq:tqr}\\
\RGR&=\frac{\Delta\NLL_{\mathrm{base}}-\Delta\NLL_{\mathrm{corrected}}}{\Delta\NLL_{\mathrm{base}}}.\label{eq:rgr}
\end{align}
TQR normalizes by the target's advantage over the source; corrected E002 TQR is 0.405755. Like NCR, it is a ratio of aggregate means, not a mean of document ratios. It is undefined for a nonpositive denominator; RGR is undefined for a nonpositive base gap. Neither normalized score is accuracy.

For distributions $p$ (native) and $q$ (condition), $m=(p+q)/2$ and
\begin{equation}
\operatorname{JS}(p,q)=\tfrac12\KL(p\|m)+\tfrac12\KL(q\|m).
\end{equation}
Natural logarithms are used. Top-5 overlap is $|\operatorname{top5}(p)\cap\operatorname{top5}(q)|/5$, and entropy is $-\sum_v q_v\log q_v$. These metrics average over the shared target positions and documents. State error is $\|S_c-S_{\rm native}\|_F/\max(\|S_{\rm native}\|_F,10^{-12})$; cosine uses flattened tensors. New attention entries are compared only after handoff.

E001's improvement fraction for document $d$ is
\begin{equation*}
f_d=\frac{\Delta\NLL_{d,\mathrm{KV}}-\Delta\NLL_{d,\mathrm{full}}}{\Delta\NLL_{d,\mathrm{KV}}}
\end{equation*}
when the denominator is positive. All 64 qualify. Its 79.7\% mean is therefore not the ratio of the aggregate 0.7473 improvement to the aggregate KV-only gap. One E001 document has a nonpositive source-minus-native denominator and is excluded from document-level TQR; the reported aggregate TQR uses all documents' mean NLLs and remains defined.

The bootstrap samples documents with replacement, preserving each document's condition pairing. Ten thousand means are drawn with NumPy's default generator using frozen seeds 2026083103 (E001) and 2026090103 (E002); percentile endpoints are 0.025 and 0.975. The RGR interval recomputes its ratio of means inside each bootstrap resample. No token-level resampling, alternate confidence procedure, or multiplicity-adjusted exploratory claim is substituted for the frozen protocol.

The E002 full-state gate requires a positive base-minus-corrected CI, remaining-gap reduction at least 0.25, corrected excess NLL at most 0.10, NCR at least 0.90, and upper CI endpoints below zero for corrected-minus-source and corrected-minus-shuffled. The stronger near-native gate additionally requires excess NLL at most 0.05, NCR at least 0.95, and top-1 agreement at least 0.90. Only that stronger outcome unlocks the 16K branch. E002's recorded verdict is \code{FULL\_STATE\_HANDOFF}, not \code{NEAR\_NATIVE\_HANDOFF}; all three stronger thresholds fail. E001's recorded verdict is \code{RECURRENT\_STATE\_TRANSLATABLE}: its mean and median document improvement fractions exceed the 25\% contribution criterion. Its fully translated condition fails the stronger full-state gate, with excess NLL 0.207032 above 0.20 and TQR $-0.413227$ below 0.75. Gates are experiment-specific. Both canonical results explicitly record the 16K branch as unrun.

\section{Full-Precision Continuation Results}
\label{app:fullresults}
\begin{figure*}[t]
\centering\includegraphics[width=\textwidth]{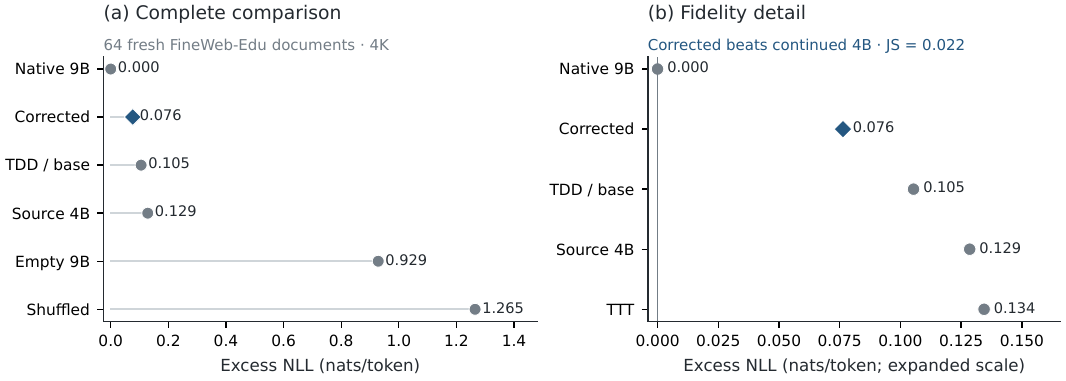}
\caption{E002's complete test comparison. Markers are means over 64 paired documents. The left axis includes empty and shuffled controls; the right expands the scale and includes TTT. Corrected excess NLL is 0.076367. Base-minus-corrected improvement is 0.029022 nats/token (95\% paired document bootstrap CI [0.022828, 0.035413]); plotted means are not independent token-level replicates.}
\label{fig:e002}
\end{figure*}

Tables~\ref{tab:e001full}--\ref{tab:e002full} retain the six-decimal numerical presentation. The included verified JSON retains full floating-point values. Figure~\ref{fig:e002} shows the Experiment 2 controls on a common overview scale and an explicitly expanded fidelity scale. No results are pooled across the two corpora.

\begin{table}[t]
\centering\footnotesize\setlength{\tabcolsep}{3pt}\begin{tabular}{lrrr}
\toprule
Condition & NLL & $\Delta$NLL & Top-1 \\
\midrule
Native 9B & 2.160916 & 0.000000 & 1.000000 \\
Source 4B & 2.307412 & 0.146496 & -- \\
Empty 9B & 3.245654 & 1.084738 & 0.581299 \\
KV only & 3.115271 & 0.954355 & 0.600098 \\
KV + direct GDN & 2.291314 & 0.130398 & 0.821289 \\
Full translated & 2.367948 & 0.207032 & 0.783203 \\
Full shuffled & 3.316181 & 1.155265 & 0.555664 \\
\bottomrule
\end{tabular}

\caption{E001 LOCKED results: 64 PG19 documents, 4K prefix, 64 teacher-forced targets. NLL and excess NLL are nats/token; top-1 is native-target prediction agreement.}
\label{tab:e001full}
\end{table}

\begin{table*}[t]
\centering\small\setlength{\tabcolsep}{7pt}\begin{tabular}{lrrrrrr}
\toprule
Condition & NLL & $\Delta$NLL & NCR & TQR & Top-1 & JS \\
\midrule
Native 9B & 1.913032 & 0.000000 & 1.000000 & 1.000000 & 1.000000 & 0.000000 \\
Source 4B & 2.041544 & 0.128512 & 0.861597 & 0.000000 & 0.833984 & 0.038052 \\
Empty 9B & 2.841564 & 0.928532 & 0.000000 & -6.225273 & 0.626465 & 0.156773 \\
TDD / base & 2.018422 & 0.105390 & 0.886498 & 0.179921 & 0.846436 & 0.028583 \\
TTT & 2.047468 & 0.134437 & 0.855216 & -0.046104 & 0.832031 & 0.036172 \\
Joint corrected & 1.989399 & 0.076367 & 0.917755 & 0.405755 & 0.864258 & 0.022067 \\
Joint shuffled & 3.177764 & 1.264732 & -0.362078 & -8.841385 & 0.578369 & 0.192132 \\
\bottomrule
\end{tabular}

\caption{E002 LOCKED results: 64 fresh FineWeb-Edu documents, 4K prefix, 64 teacher-forced targets. All entries are document means or defined ratios of document means. NLL and excess NLL use nats/token. NCR, TQR, top-1 agreement, and JS have distinct meanings; native/empty NCR and native/source TQR are fixed by definition.}
\label{tab:e002full}
\end{table*}

\section{Complete Component Factorial}
\label{app:factorial}
Table~\ref{tab:factorialall} includes every D/T cell. Coding D as $-1$ and T as $+1$, a main effect is the within-document signed sum over eight cells divided by four. Pairwise interactions divide by two, yielding a difference-in-differences averaged over the remaining factor; the three-way contrast is the full signed sum. Table~\ref{tab:factorialeffects} shows these contrasts and their frozen intervals. A LOCKED interaction can have an interval excluding zero and still fail the preregistered 0.02-nat materiality criterion. That is the case for $K\times R$. We therefore report failure of the specified material replication criterion, rather than asserting that all interactions are zero.

\begin{table}[t]
\centering\small\begin{tabular}{lrr}
\toprule
Cell & Validation & LOCKED \\
\midrule
DDD & 0.217430 & 0.165163 \\
DDT & 0.217473 & 0.166589 \\
DTD & 0.231225 & 0.178138 \\
DTT & 0.231250 & 0.180053 \\
TDD & 0.133377 & 0.105390 \\
TDT & 0.132226 & 0.105677 \\
TTD & 0.175278 & 0.133202 \\
TTT & 0.174644 & 0.134437 \\
\bottomrule
\end{tabular}

\caption{Every E002 factorial cell's excess NLL, in nats/token. Validation has 32 fresh FineWeb-Edu documents; LOCKED has 64 different documents. Both use 4K prefixes. Base selection uses validation alone.}
\label{tab:factorialall}
\end{table}

\begin{table*}[t]
\centering\small\setlength{\tabcolsep}{7pt}\begin{tabular}{lrrrr}
\toprule
Contrast & Validation & 95\% CI & LOCKED & 95\% CI \\
\midrule
KV & -0.070463 & [-0.100867, -0.040946] & -0.052809 & [-0.069557, -0.037096] \\
Recurrent & 0.027973 & [0.010360, 0.047548] & 0.020753 & [0.007905, 0.034039] \\
Convolution & -0.000429 & [-0.002647, 0.001774] & 0.001216 & [-0.000055, 0.002580] \\
$K\times R$ & 0.028373 & [0.008556, 0.057582] & 0.015067 & [0.008212, 0.022371] \\
$K\times C$ & -0.000927 & [-0.002873, 0.001039] & -0.000910 & [-0.002206, 0.000378] \\
$R\times C$ & 0.000250 & [-0.002447, 0.003218] & 0.000718 & [-0.001490, 0.003543] \\
$K\times R\times C$ & 0.000535 & [-0.003747, 0.005266] & 0.000458 & [-0.001807, 0.002798] \\
\bottomrule
\end{tabular}

\caption{E002 orthogonal within-document factorial contrasts, in nats/token, with 95\% paired document bootstrap intervals. Main effects are translated minus direct, averaged over the other two components. Pairwise material replication requires matching signs and absolute LOCKED estimate at least 0.02; no pairwise contrast satisfies both requirements.}
\label{tab:factorialeffects}
\end{table*}

\section{Secondary State Diagnostics and Post-Verdict Ablations}
\label{app:diagnostics}
\begin{figure*}[t]
\centering\includegraphics[width=\textwidth]{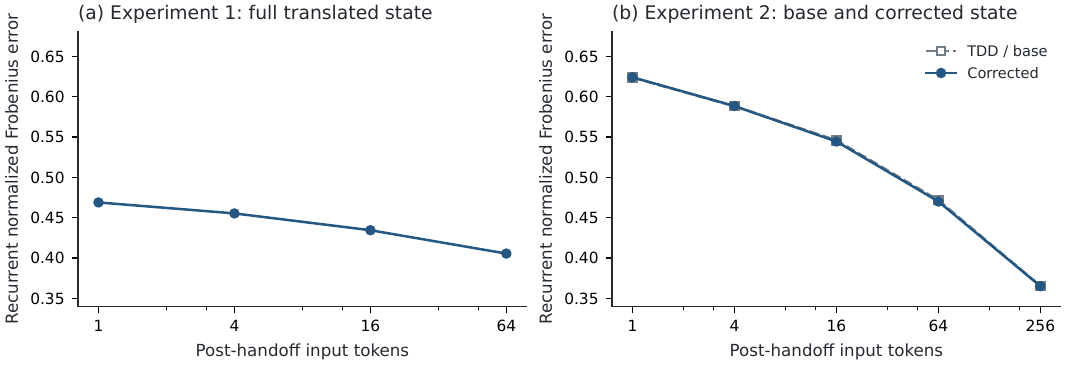}
\caption{Secondary recurrent-state convergence under shared teacher-forced inputs. Each marker averages layer/head state errors over 64 documents within the indicated experiment. The horizontal axis is logarithmic with only the recorded checkpoints shown. E001 and E002 use different LOCKED corpora and are not paired with one another. E002's 256-input state measurement is not a 16K-context test or a free-generation evaluation.}
\label{fig:repair}
\end{figure*}

Figure~\ref{fig:repair} presents the preregistered state-tracking endpoints from the frozen evidence. E001 statistics are summarized in a post-verdict derived artifact, but the tracked checkpoints and measurements were specified before the LOCKED run. The E002 base/corrected trajectories are close, and corrected state is not uniformly closer at the earliest inputs. Continued target computation reduces recurrent mismatch without eliminating it. Neither state distance nor cosine is a success gate. E001 full-translated recurrent normalized error declines from 0.4687 after one new input to 0.4053 after 64. E002 corrected error declines from 0.6239 to 0.3650 through 256 inputs, while cosine rises from 0.7778 to 0.9266. Shared teacher inputs and gated forgetting can reduce discrepancy without recovering all prefix information.

\subsection{Post-verdict component removals}
\label{app:ablations}
After the 4K verdict was frozen, E002 disabled components of the selected correction on the same LOCKED documents. Table~\ref{tab:ablations} shows the resulting impact relative to the full correction. Removing recurrent correction has the largest aggregate effect among complete-component removals; removing convolution correction has a small interval spanning zero. The middle recurrent third is the most sensitive of the three depth blocks under this intervention. These are post-verdict diagnostics with no multiplicity correction, no refitting, and no independent confirmation set. They describe sensitivity of the fitted correction; they do not establish the best sparse architecture or authorize changing the selected result.

\begin{table*}[t]
\begin{minipage}[t]{0.485\textwidth}
\centering\footnotesize\setlength{\tabcolsep}{2pt}\begin{tabular}{lrrr}
\toprule
Removed & $\Delta$NLL & Impact & 95\% CI \\
\midrule
Convolution & 0.077026 & 0.000658 & [-0.000387, 0.001685] \\
R: early third & 0.082034 & 0.005666 & [0.002873, 0.008594] \\
KV & 0.083648 & 0.007281 & [0.004475, 0.010285] \\
R: late third & 0.077912 & 0.001545 & [0.000136, 0.002925] \\
R: middle third & 0.086258 & 0.009891 & [0.006875, 0.013005] \\
R: all layers & 0.095287 & 0.018920 & [0.014069, 0.024019] \\
\bottomrule
\end{tabular}

\caption{Post-verdict E002 ablations on the same 64 LOCKED documents. R denotes recurrent state. The named correction is removed; positive impact is increased NLL relative to the complete correction. All values are nats/token; intervals are paired document bootstrap intervals. These diagnostics did not determine or revise the canonical verdict.}
\label{tab:ablations}
\end{minipage}
\hfill
\begin{minipage}[t]{0.485\textwidth}
\centering\footnotesize\setlength{\tabcolsep}{4pt}\begin{tabular}{lr}
\toprule
Recorded stage & Median (ms) \\
\midrule
Native 9B prefill & 886.646 \\
Source 4B prefill & 634.911 \\
Base construction & 0.041 \\
Correction & 69.145 \\
State installation & 15.829 \\
Bridge token & 35.410 \\
Difference of prefill medians & 251.736 \\
Reported handoff margin & 131.310 \\
\bottomrule
\end{tabular}

\caption{E002's frozen engineering timing summary. Primitive values are document wall-clock medians; the final two rows are arithmetic combinations of those medians. The accounting excludes the preceding state-translation pass and other costs discussed in the text, so the reported margin is not a measured production speedup.}
\label{tab:timing}
\end{minipage}
\end{table*}

\section{Prototype Timing and Its Accounting Boundary}
\label{app:timing}
Table~\ref{tab:timing} preserves E002's recorded timing summary. Each primitive is aggregated as its document median. The available budget is the native-prefill median minus the source-prefill median; the reported margin further subtracts the sum of base construction, correction, installation, and bridge medians. A difference or sum of medians is not an independently measured median end-to-end latency.

The code makes an additional exclusion explicit: base-state construction selects components from \emph{already materialized} source and translated states. The translation pass occurs earlier and is not charged to the reported 131.310-ms margin. This margin also does not establish the full cost of extraction, serialized evidence I/O, model loading, batching, scheduling, or a GPU-local deployment. The original report's ``end-to-end'' characterization is therefore too broad for the underlying aggregation; the numeric artifact is retained, and the narrower code-defined accounting is reported here. No production latency saving follows from the table.

\section{Evidence Integrity and Paper Reproduction}
\label{app:reproduction}
The paper uses the frozen \code{e001\_handoff} and \code{e002\_coupler} directories supplied with the specification. The LOCKED raw evidence and E002 fitting artifacts are stored in the corresponding original E001/E002 archive directories. Their raw SHA-256 values match the canonical result references:
\begin{quote}\footnotesize
E001: \path{1639038217b9bb376de33be57ab624e083a093a45dc5a5f8749b565247c9e546}\par
E002: \path{a189c454a4f52a8ceb03ec09e0e109efa139ce01225a472c59a70148acd0c629}
\end{quote}
E001's frozen execution manifest hash is
\begin{quote}\footnotesize\path{69bcca1bea1ecce2cc5c7c9728581aeca8ed9abf56028c2cd9c2fd7498efb0c9},\end{quote}
and E002's is
\begin{quote}\footnotesize\path{4ea9e85a2450a6c0f8dbb979f6ecfba32d4f45606103a946a32595208359070c}.\end{quote}
The selected correction tensor SHA-256 is
\begin{quote}\footnotesize\path{3565d5ddb81b9c195e0cc02e143acf85a49f8fe732a794fe5d1d1c2463cf6331}.\end{quote}
The combined metadata-plus-tensor correction hash is distinct from this file hash:
\begin{quote}\footnotesize\path{e93a15a21b6eba37ec794d94c45780059a3653f9466c26db1b20960419260db3}.\end{quote}

The accompanying LaTeX project contains a claim ledger, an evidence map, input hashes, verification outputs, and deterministic figure/table scripts. The extraction script reads frozen raw records and reproduces aggregate NLLs, paired bootstrap intervals, factorial selection and contrasts, normalized metrics, and correction inventory. It performs no model evaluation, translator fitting, correction training, or LONG evaluation. Rendering scripts operate on that verified extraction and write only into the paper project. All reported primary intervals remain the canonical frozen intervals. The README gives exact local source paths and build commands.

\end{document}